\documentclass{article}
\usepackage{spconf,amsmath,graphicx}
\usepackage{enumitem}
\setlist{nosep, leftmargin=14pt}
\usepackage{xcolor,booktabs,multirow}
\usepackage{epstopdf}
\usepackage{tikz}
\usetikzlibrary{arrows.meta,positioning}
\usepackage[font=small,labelfont=bf,skip=2pt]{caption}
\usepackage{url}
\title{Domain-Specific Self-Supervised Representation Learning for Retinal Fundus Classification}

\name{Bekzat Nurlanbekova$^{\star}$ \qquad Fung Fung Ting$^{\dagger}$ }
\address{$^{\star \dagger}$ School of Information Technology, Monash University, Malaysia Campus, Subang Jaya, Malaysia \\
${\dagger}$Corresponding author. Email: ting.fung@monash.edu}

\begin{document}
\maketitle

\begin{abstract}
Annotated medical images remain scarce and costly to obtain, which motivates self-supervised learning (SSL). We present a controlled empirical study of two SSL frameworks, SimSiam and SimCLR, for retinal fundus classification, asking whether retina-specific augmentations (Graham's local-mean subtraction and CLAHE) improve representations beyond standard augmentation and grayscale conversion. ResNet-18 backbones are pretrained on 48K unlabeled fundus images and evaluated by linear probing and fine-tuning on EDID (multi-disease) and RetinaMNIST (DR grading). On EDID, grayscale conversion accounts for most of SimSiam's linear-probe gain, and Graham or CLAHE add little on top of it. After fine-tuning, SimSiam with Graham+Gray exceeds an ImageNet-supervised baseline on RetinaMNIST (+3.0 accuracy points) but only matches it on EDID. Retina-specific augmentation therefore helps inconsistently and should be validated per SSL method and task.
\end{abstract}


\section{Introduction}
\label{sec:intro}
CNNs achieve strong performance in classification, detection and segmentation \cite{rakshitcnn}, including medical tasks such as detecting pneumonia, brain tumors and eye diseases \cite{chenReviewConvolutionalNeural2025}. Supervised training, however, needs large labeled datasets, and medical annotation is costly, slow and dependent on specialists.

Self-supervised learning (SSL) reduces this dependence. SimCLR \cite{chenSimpleFrameworkContrastive2020}, a contrastive framework, has shown promise in histopathology, dermatology and chest X-ray classification \cite{bioengineering13020176, jaw_hist2025}. SimSiam \cite{Chen_2021_CVPR}, a non-contrastive framework, has been applied to breast cancer, histopathology and chest X-ray analysis \cite{diagnostics16091319, voigtInvestigationSemiSelfsupervised2023, huang2024systematic}, yet it is less studied in medical imaging than contrastive methods.

SSL learns from multiple augmented views of the same image, so the choice of augmentation is critical. Stacke et al.~\cite{stackeLearningRepresentationsContrastive2022} showed that its effect on SimCLR is dataset- and task-dependent in histopathology. In retinal imaging, Graham's preprocessing and contrast-limited adaptive histogram equalization (CLAHE) benefit supervised classification \cite{macsikImagePreprocessingbasedEnsemble2024, kassaniDiabeticRetinopathyClassification2019, zotero-item-240}, but their effect on SSL representations across several retinal diseases is underexplored.

We therefore study SimCLR and SimSiam with a ResNet-18 backbone under a three-stage protocol (SSL pretraining on unlabeled fundus images, linear evaluation, fine-tuning) and ask: (i) how the two frameworks compare across training configurations, (ii) whether retina-specific augmentations improve on standard augmentation under limited pretraining, and (iii) whether the gains transfer across retinal tasks.

\section{Methods}
\label{sec:methods}
\subsection{Self-supervised frameworks}
SimCLR \cite{chenSimpleFrameworkContrastive2020} pulls together two augmented views of the same image and pushes apart all other samples in the batch ($2N{-}2$ negatives for batch size $N$), using the NT-Xent loss with temperature $\tau$ on L2-normalized projections. SimSiam \cite{Chen_2021_CVPR} uses no negatives: a shared encoder processes both views, a predictor is added to one branch and a stop-gradient to the other, and the loss maximizes their cosine similarity. This asymmetry prevents collapse without large batches, so the two frameworks represent complementary contrastive and negative-free paradigms.

\subsection{Augmentation strategies}
Retinal findings such as drusen, microaneurysms and an enlarged cup-to-disc ratio are subtle, so aggressive color, blur or crop transformations may remove diagnostic cues. We therefore add retina-specific operations to a standard pipeline (horizontal flip, $\pm15^\circ$ rotation, random resized crop with scale 0.2--1.0, color jitter, Gaussian blur): random grayscale (Gray), CLAHE, and Graham's preprocessing, each applied with $p{=}0.1$. \emph{Combined} applies all three. Preprocessing is used only during SSL pretraining; fine-tuning and inference use raw images.  

\begin{figure}[t]
\centering
\begin{tabular}{cc}
\begin{tabular}{c}
\begin{tikzpicture}[
    node distance=3.5mm and 4.5mm,
    box/.style={draw, rounded corners=1pt, align=center, font=\scriptsize, inner sep=2pt, minimum height=6mm},
    arr/.style={-{Stealth[length=1.6mm]}, thick}]
\node[box] (d) {FISSL\\[-1pt]48K unlabeled};
\node[box, below=of d] (a) {Augment\\[-1pt]standard $+$ Gray/\\[-1pt]Graham/CLAHE};
\node[box, below=of a] (s) {SimSiam/SimCLR\\[-1pt]ResNet-18};
\node[box, below=of s, fill=green!20] (f) {Fine-tune\\[-1pt]or frozen probe};
\node[box, below=of f] (r) {Raw images\\[-1pt]EDID, RetinaMNIST};
\draw[arr] (d) -- (a); 
\draw[arr] (a) -- (s); 
\draw[arr] (s) -- (f); 
\draw[arr] (r) -- (f);
\node[font=\scriptsize\bfseries, anchor=south west] at (d.north west) {(a)};
\end{tikzpicture}
\end{tabular} 
& 
\begin{tabular}{c}
\textbf{\scriptsize (b)}\\[2pt]
\includegraphics[width=0.56\linewidth]{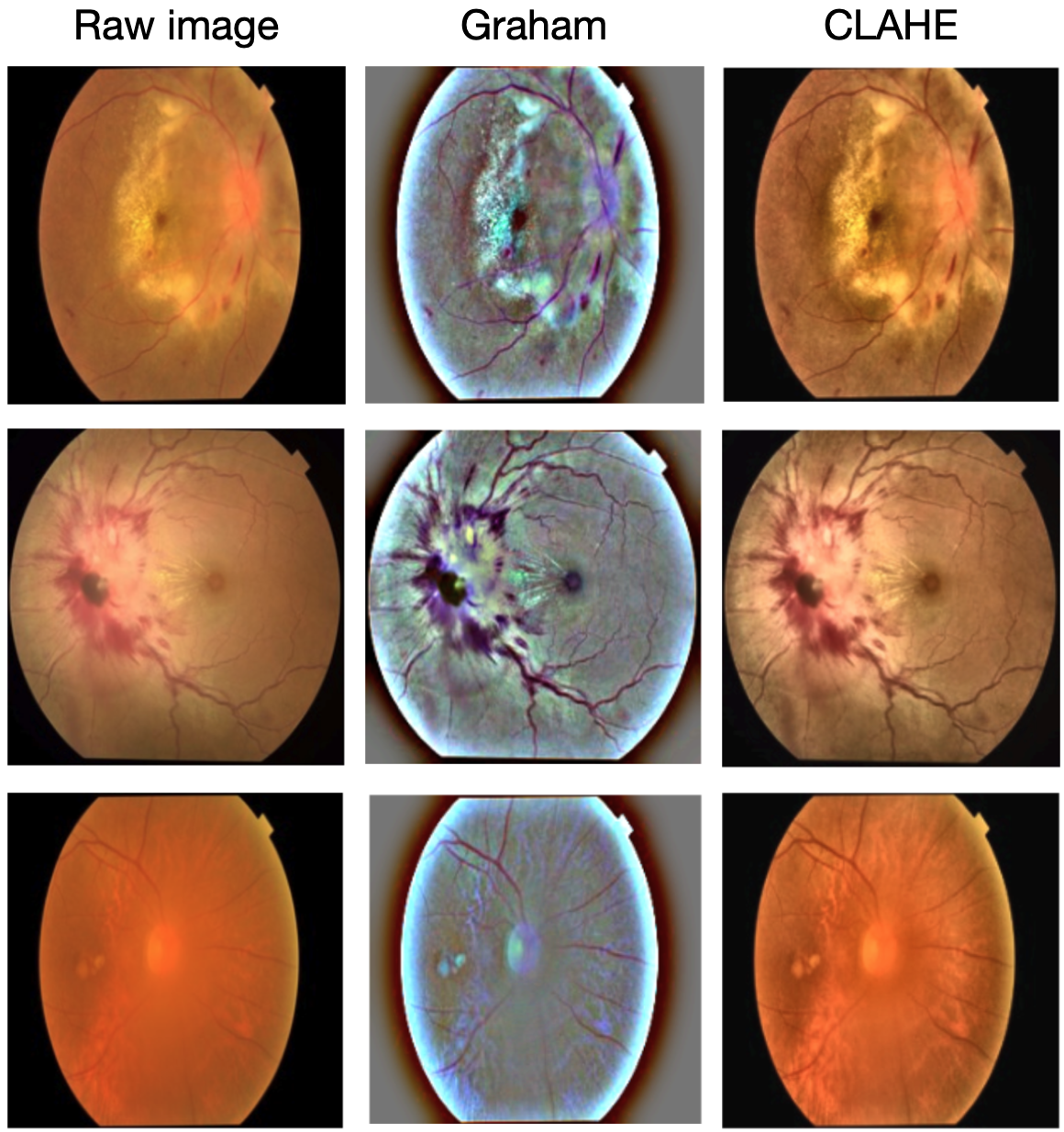}
\end{tabular}
\end{tabular}
\caption{(a) Study pipeline. (b) Original fundus images and their Graham- and CLAHE-preprocessed versions.}
\label{fig:pipeline}
\end{figure}

\textbf{Graham's preprocessing} \cite{zotero-item-240} subtracts a Gaussian-smoothed copy of the image from the original, then rescales and shifts the result by a constant $\gamma$. It removes low-frequency illumination and accentuates local structure such as vessels and lesions (Fig.~\ref{fig:pipeline}b); it is a form of high-boost (unsharp-mask-style) filtering; with $\sigma$=10, $\gamma$=128. \textbf{CLAHE} \cite{zuiderveld1994} equalizes histograms within small tiles and bounds amplification with a clip limit = 2.0 and tile size=(8, 8). Unlike Graham's method, it enhances local contrast without correcting illumination gradients.

\subsection{Datasets}
\textbf{FISSL} \cite{fissl-huggingface} provides about 48K unlabeled $224{\times}224$ fundus images compiled from RFMiD, ODIR, EyePACS and APTOS for SSL pretraining. \textbf{EDID} \cite{sharminDatasetColorFundus2024} contains 5,335 fundus photographs for multi-class classification, split 70/10/20\% (train/validation/test, stratified). It is highly imbalanced: diabetic retinopathy, glaucoma and healthy each exceed 1,000 images, macular scar and myopia have about 400--500, and CSCR, disc edema, retinal detachment and retinitis pigmentosa about 100 each. \textbf{RetinaMNIST} \cite{yangMedMNISTV2Largescale2023} (MedMNIST v2) has 1,600 $224{\times}224$ images graded into five diabetic retinopathy stages, split 1,080/120/400 for train/validation/test.

\subsection{Training and evaluation}
All experiments use ResNet-18 \cite{heDeepResidualLearning2015}. SSL models are pretrained for 200 epochs (10-epoch linear warm-up, checkpoints every 50 epochs) with batch sizes 128 and 256. \emph{SimSiam}: SGD (momentum 0.9, weight decay $10^{-4}$), a 3-layer projection and 2-layer prediction MLP with 2048-d outputs \cite{Chen_2021_CVPR}, learning rate $0.05\times\text{batch}/256$ with cosine decay. \emph{SimCLR}: LARS (momentum 0.9, weight decay $10^{-6}$), learning rate $0.3\times\text{batch}/256$ \cite{chenSimpleFrameworkContrastive2020}, $\tau{=}0.5$.

\emph{Linear evaluation} trains only the final FC layer on the frozen backbone (100 epochs, batch 32, SGD with momentum 0.9 and no weight decay, learning rate 0.0125, cosine schedule). \emph{Fine-tuning} updates all layers (learning rate $10^{-3}$, 5-epoch warm-up then cosine schedule, SGD with momentum 0.9 and weight decay $10^{-4}$, up to 50 epochs with early stopping). A class-balanced weighted sampler counters label imbalance. The supervised baseline is an ImageNet-initialized ResNet-18 trained on the labeled data. Downstream training uses five random seeds, and we report mean $\pm$ SD of accuracy and macro-F1 (EDID) and accuracy and quadratic weighted kappa, QWK (RetinaMNIST).

\section{Results and Discussion}
\label{sec:results}
\begin{figure*}[t]
\centering
\includegraphics[width=0.325\textwidth]{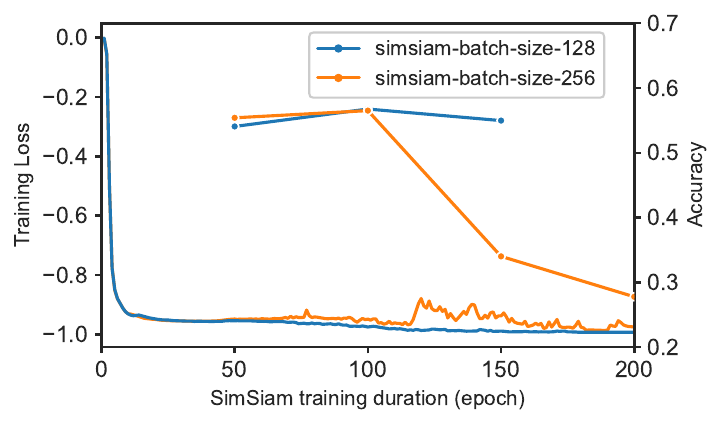}\hfill
\includegraphics[width=0.325\textwidth]{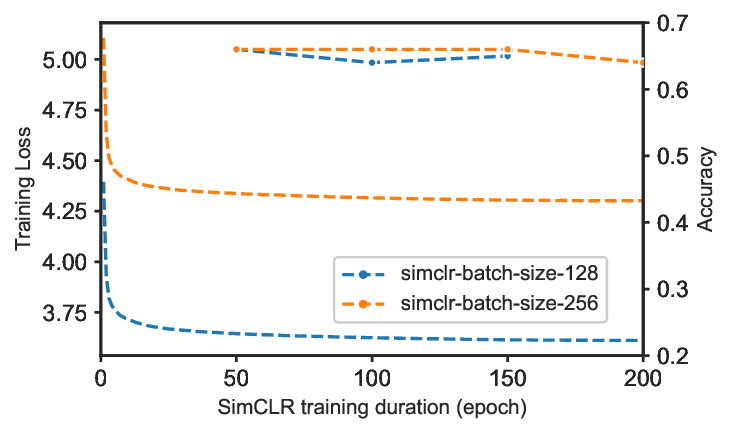}\hfill
\includegraphics[width=0.325\textwidth]{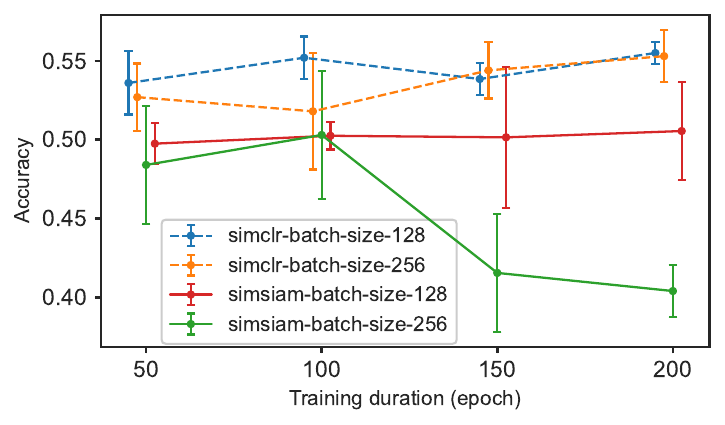}
\caption{Standard augmentation, batch sizes 128 and 256. Left, middle: SSL training loss (left axis) and EDID linear-evaluation accuracy (right axis) for SimSiam and SimCLR. Right: RetinaMNIST linear-evaluation accuracy at 50--200 pretraining epochs; error bars show SD.}
\label{fig:duration}
\end{figure*}

\subsection{Training duration and batch size}
Under standard augmentation, a continued decrease in SSL loss does not improve downstream accuracy (Fig.~\ref{fig:duration}, left and middle): pretraining beyond early convergence gives little benefit, in line with SimCLR on histology \cite{stackeLearningRepresentationsContrastive2022}. SimSiam is the more sensitive framework. At batch size 256 its linear accuracy declines after 100 epochs on both EDID and RetinaMNIST (Fig.~\ref{fig:duration}, left and right), whereas SimCLR is stable across batch sizes and durations. This loss-accuracy divergence suggests that SimSiam representations may lose diversity over time, and that standard augmentation may not give it sufficiently informative positive pairs for retinal images.

\begin{table}[t]
\centering
\fontsize{7}{8}\selectfont
\setlength{\tabcolsep}{2.6pt}
\caption{Linear evaluation (a; 50 pretraining epochs, batch size 256) and fine-tuning (b) on EDID and RetinaMNIST, mean $\pm$ SD over five seeds. Combined = CLAHE + Graham + Gray. Bold = best per column within each block; differences within one SD should not be over-interpreted.}
\label{tab:main}
\resizebox{\columnwidth}{!}{%
\begin{tabular}{lcccc}
\toprule
 & \multicolumn{2}{c}{EDID} & \multicolumn{2}{c}{RetinaMNIST} \\
\cmidrule(lr){2-3}\cmidrule(lr){4-5}
Method & Acc (\%) & Macro-F1 (\%) & Acc (\%) & QWK (\%) \\
\midrule
\multicolumn{5}{l}{\textit{(a) Linear evaluation: SimSiam}} \\
Standard      & 55.40$\pm$1.53 & 47.18$\pm$1.35 & 48.40$\pm$1.80 & 58.05$\pm$2.92 \\
+Gray         & 65.70$\pm$1.35 & 59.39$\pm$1.78 & \textbf{56.60}$\pm$0.30 & 65.07$\pm$0.86 \\
+Graham(p=0.1)       & 64.87$\pm$1.29 & 55.24$\pm$2.12 & 52.00$\pm$2.40 & 63.10$\pm$1.36 \\
+Graham(p=0.2) & 57.52$\pm$2.24 & 49.43$\pm$2.36 & 50.75$\pm$2.67 & 60.89$\pm$2.91 \\
+Graham+Gray  & \textbf{65.73}$\pm$0.78 & \textbf{59.89}$\pm$1.10 & 53.95$\pm$1.32 & \textbf{71.02}$\pm$0.59 \\
+CLAHE        & 56.79$\pm$1.95 & 50.49$\pm$3.05 & 49.60$\pm$2.45 & 58.57$\pm$3.74 \\
+CLAHE+Gray   & 65.08$\pm$2.63 & 59.04$\pm$2.58 & 51.65$\pm$3.19 & 62.77$\pm$1.33 \\
+Combined     & 62.20$\pm$1.29 & 55.73$\pm$0.87 & 52.50$\pm$2.47 & 66.11$\pm$1.76 \\
\midrule
\multicolumn{5}{l}{\textit{(a) Linear evaluation: SimCLR}} \\
Standard      & 66.13$\pm$0.45 & 58.78$\pm$0.85 & 52.70$\pm$1.12 & 62.80$\pm$2.01 \\
+Gray         & 64.53$\pm$1.36 & 59.39$\pm$1.55 & 47.25$\pm$2.71 & 63.00$\pm$1.32 \\
+Graham(p=0.1)       & 64.02$\pm$0.97 & 56.09$\pm$1.53 & 49.70$\pm$2.55 & 61.36$\pm$2.50 \\
+Graham(p=0.2) & 62.86$\pm$1.75 & 55.38$\pm$1.36 & \textbf{53.10}$\pm$2.96 & 63.58$\pm$1.65 \\
+Graham+Gray  & \textbf{67.11}$\pm$0.91 & \textbf{61.55}$\pm$0.57 & 52.95$\pm$1.49 & \textbf{65.94}$\pm$1.39 \\
+CLAHE        & 64.34$\pm$1.62 & 56.72$\pm$2.18 & 50.95$\pm$2.90 & 60.08$\pm$1.63 \\
+CLAHE+Gray   & 63.06$\pm$1.60 & 55.19$\pm$2.36 & 49.95$\pm$1.54 & 62.93$\pm$2.31 \\
+Combined     & 65.51$\pm$1.52 & 60.26$\pm$0.77 & 50.10$\pm$2.13 & 63.86$\pm$2.07 \\
\midrule
\multicolumn{5}{l}{\textit{(b) Fine-tuning}} \\
Random Init.           & 56.79$\pm$3.12 &  50.75$\pm$3.19 & 48.05$\pm$2.46 & 52.97$\pm$3.03\\
ImageNet supervised    & 75.75$\pm$0.38 & \textbf{75.80}$\pm$0.40 & 60.08$\pm$1.06 & 72.27$\pm$1.37 \\
SimSiam +Standard      & 74.94$\pm$1.77 & 72.81$\pm$2.07 & 55.25$\pm$1.90 & 73.70$\pm$2.02\\
SimCLR +Standard      & 75.20$\pm$1.95 & 72.81$\pm$1.31  & 57.20$\pm$1.91 & 74.49$\pm$0.99 \\

SimSiam +Gray          & 74.94$\pm$1.25 & 74.53$\pm$0.98 & 54.30$\pm$4.54 &72.41$\pm$3.42\\
SimCLR +Gray           & 72.82$\pm$0.98 & 71.80$\pm$1.41  & 59.35$\pm$1.31 & 74.76$\pm$1.92 \\
SimSiam +Graham+Gray   & \textbf{75.88}$\pm$0.92 & 75.15$\pm$0.62 & \textbf{63.08}$\pm$0.94 & \textbf{76.06}$\pm$1.28 \\
SimCLR +Graham+Gray    & 73.81$\pm$0.90 & 72.48$\pm$0.76 & 57.42$\pm$1.48 & 74.01$\pm$0.54 \\
SimSiam +CLAHE+Gray    & 73.01$\pm$1.73 & 70.97$\pm$1.99 & 60.25$\pm$2.13 & 75.65$\pm$2.14 \\
SimCLR +CLAHE+Gray     & 72.65$\pm$2.72 & 71.51$\pm$2.64 & 59.75$\pm$1.83 & 74.75$\pm$2.38 \\
\bottomrule
\end{tabular}%
}
\end{table}

\begin{figure}[t]
\centering
\includegraphics[width=\columnwidth]{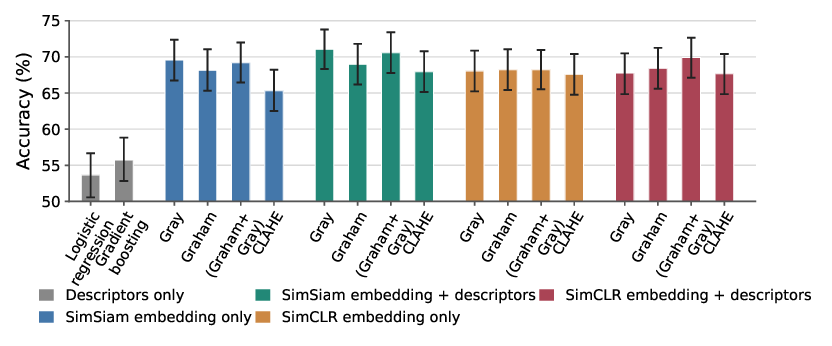}
\caption{EDID classification of Frangi-derived descriptors, frozen SSL embeddings, and their concatenation with logistic regression. Error bars: 95\% bootstrap CI. }
\label{fig:descriptor}
\end{figure}

\subsection{Linear evaluation and Fine-tuning}
Table~\ref{tab:main}(a) reports linear evaluation of the 50-epoch, batch-256 models. For SimSiam on EDID, grayscale conversion is the main driver: +Gray raises accuracy from 55.40 to 65.70; Graham+Gray (65.73) is indistinguishable from +Gray; CLAHE behaves similarly (+CLAHE+Gray 65.08). On RetinaMNIST, +Gray gives the best SimSiam accuracy (56.60), whereas Graham+Gray gives the best QWK (71.02 vs.\ 65.07). SimCLR shows no grayscale benefit: Standard is already strong on EDID (66.13), and Graham+Gray adds 0.98 points (about one SD). On RetinaMNIST, SimCLR variants mostly differ within their SDs, except that +Gray lowers accuracy (47.25 vs.\ 52.70). SimSiam is sensitive to augmentation, increase of Graham to p=0.2 reduced performance.

After \emph{fine-tuning} (Table~\ref{tab:main}(b)), SimSiam with Graham+Gray reaches the highest RetinaMNIST accuracy (63.08) and QWK (76.06), improving on the ImageNet-supervised baseline by 3.00 and 3.79 points. On EDID it matches the baseline in accuracy (75.88 vs.\ 75.75) but has lower macro-F1 (75.15 vs.\ 75.80). SimCLR generally trails SimSiam, and CLAHE+Gray does not improve on Graham+Gray, although every SSL variant exceeds the baseline's RetinaMNIST QWK.

\begin{figure}[t]
\centering
\scriptsize
\begin{tabular}{@{}p{0.18\columnwidth}p{0.18\columnwidth}p{0.18\columnwidth}p{0.18\columnwidth}@{}}
\hspace{25pt}Raw & \hspace{10pt}Supervised & \hspace{4pt}SimSiam & \hspace{3pt}SimCLR
\end{tabular}\\[-1pt]
\includegraphics[width=0.75\columnwidth]{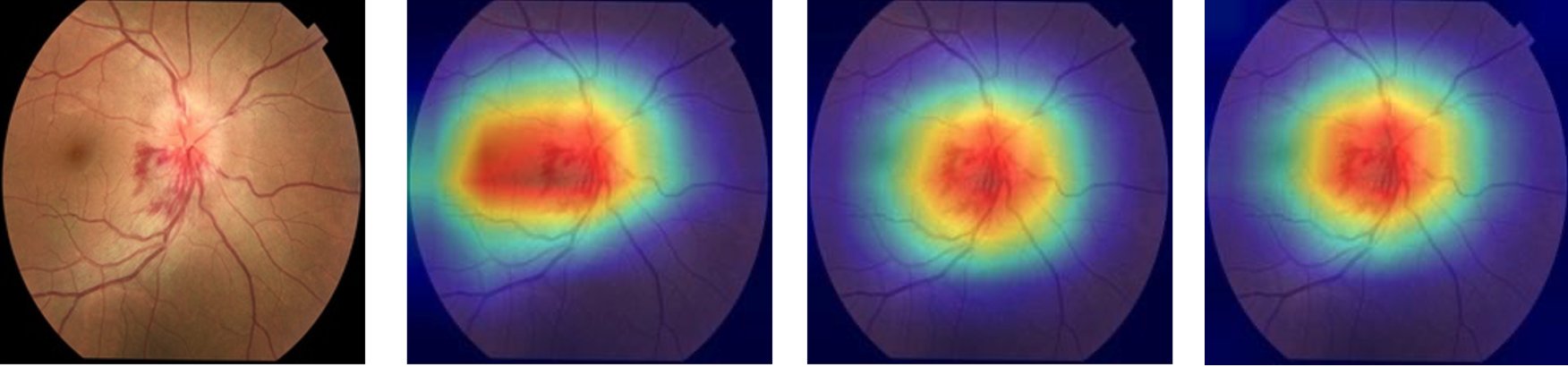}\\[-1pt]
\includegraphics[width=0.75\columnwidth]{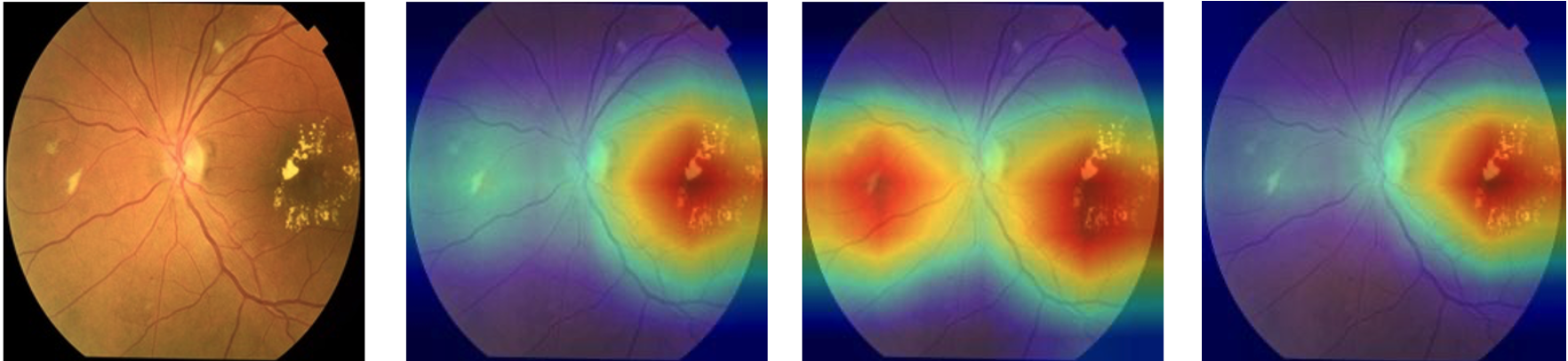}\\[-1pt]
\includegraphics[width=0.75\columnwidth]{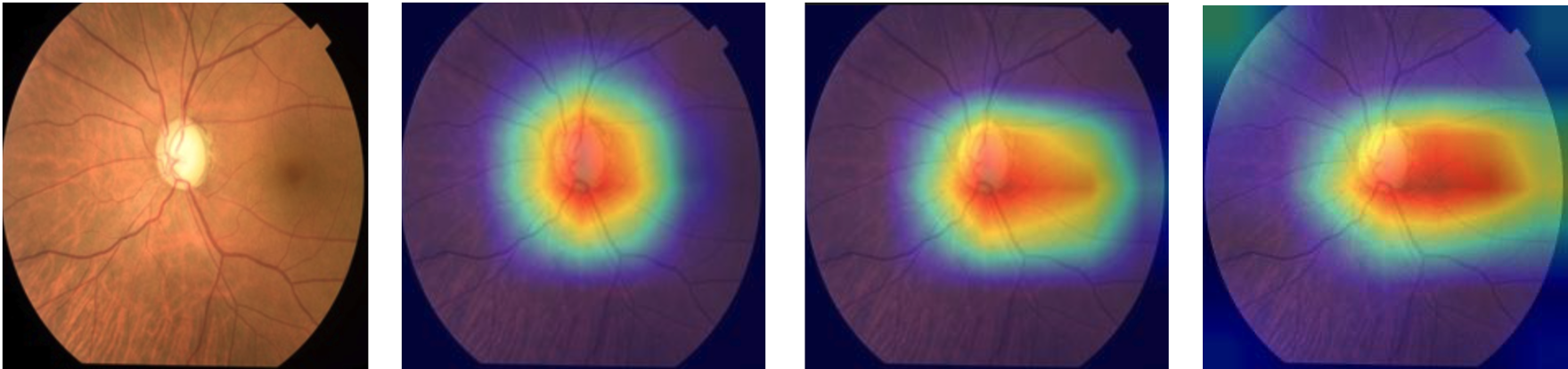}
\caption{Grad-CAM for disc edema, DR and glaucoma (top to bottom): raw image and models with ImageNet supervision, SimSiam and SimCLR (Standard+Graham+Gray pretraining).}
\label{fig:gradcam}
\end{figure}

Grad-CAM maps (Fig.~\ref{fig:gradcam}) show both SSL models highlighting pathology-related regions in disc edema and DR, whereas in the glaucoma example attention drifts toward the macula and fovea instead of staying on the optic disc.

\subsection{Post-hoc representation analysis}

SSL pretraining may under-represent fine-grained vascular and morphometric features. For post-hoc analysis, retinal structural descriptors were handcrafted using Frangi filtering on the green channel, hysteresis thresholding (low=0.18, high=0.35), adapting the ring-based approach of Li et al.~\cite{liInterpretableFundusImage2026}. Five optic-disc-centred rings within the FOV were used to extract structural (skeleton density, area fraction, mean width, width IQR) and colour/intensity descriptors (red-green ratio, entropy difference) from raw EDID images. Frangi responses capture both vascular and non-vascular structures. As shown in Fig.~\ref{fig:descriptor}, descriptor-only baselines underperform SSL embeddings, while concatenation slightly improves accuracy by up to 2.63\% (SimSiam+CLAHE). SimSiam+Gray with descriptors achieves the highest linear probe accuracy (71.05\%).

\subsection{Discussion}
Retina-specific augmentation helped inconsistently. For SimSiam on EDID, grayscale conversion rather than Graham or CLAHE explains most of the linear-evaluation gain, and the effect of each augmentation depends on the SSL framework, the metric and the dataset. The one consistent gain over the supervised baseline, SimSiam with Graham+Gray on RetinaMNIST, cannot yet be attributed to Graham alone, because improvement over fine-tuned Standard and Gray controls was modest. This favors evaluating augmentations per method and task rather than assuming a general benefit from retina-oriented preprocessing.

\emph{Limitations.} We use one backbone, one pretraining set and two downstream tasks. FISSL draws on DR-oriented sources (EyePACS, APTOS), which may favor DR-related tasks, and the glaucoma example suggests that disc-centered cues may be under-represented. SimCLR was run with a fixed temperature and batch sizes of 128 and 256, so tuning these may narrow its gap to SimSiam.

\subsection{Compliance with Ethical Standards and Conflict of Interest}
This study used publicly available RetinaMNIST, FISSL and EDID datasets. No new data were collected from human participants.

No funding was received for conducting this study. The authors have no relevant financial or non-financial interests to disclose.

\bibliographystyle{IEEEbib}
\bibliography{references-fixed}

\end{document}